\documentclass[12pt]{article}

\usepackage[margin=1in]{geometry} 
\usepackage[utf8]{inputenc}
\usepackage[T1]{fontenc}
\usepackage{lmodern}
\usepackage{microtype}

\usepackage{amsmath, amssymb, amsthm}
\usepackage{bm}

\usepackage{graphicx}
\usepackage{booktabs}

\usepackage[nomarkers]{endfloat}

\usepackage[numbers,square,sort&compress]{natbib}
\usepackage{hyperref}
\hypersetup{
  colorlinks=true,
  linkcolor=blue,
  citecolor=blue,
  urlcolor=blue
}

\newcommand{\NN}{\mathbb{N}}
\newcommand{\E}{\mathbb{E}}
\newcommand{\EE}{\mathbb{E}}
\newcommand{\PP}{\mathbb{P}}
\newcommand{\Var}{\mathrm{Var}}
\newcommand{\vTwo}{v_2}
\newcommand{\1}{\mathbf{1}}
\newcommand{\roundodd}[1]{\left\lfloor #1 \right\rceil_{\text{odd}}}

\title{Bayesian Modeling of Collatz Stopping Times:\\
\large A Probabilistic Machine Learning Perspective}
\author{
Nicol\`o Bonacorsi\thanks{
Department of Applied Mathematics, Columbia University, New York, USA.
Email: \href{mailto:nb3328@columbia.edu}{nb3328@columbia.edu}
}
\and
Matteo Bordoni\thanks{
Department of Mathematics, Columbia University, New York, USA.
Email: \href{mailto:mb5333@columbia.edu}{mb5333@columbia.edu}
}
}
\date{\today}

\begin{document}
\maketitle

\begin{abstract}
We study the Collatz total stopping time $\tau(n)$ over $n\le 10^7$ from a probabilistic machine learning
viewpoint. Empirically, $\tau(n)$ is a skewed and heavily overdispersed count with pronounced arithmetic
heterogeneity. We develop two complementary models.
First, a Bayesian hierarchical Negative Binomial regression (NB2-GLM) predicts $\tau(n)$ from simple
covariates ($\log n$ and residue class $n \bmod 8$), quantifying uncertainty via posterior and posterior
predictive distributions.
Second, we propose a mechanistic generative approximation based on the odd-block decomposition: for odd
$m$, write $3m+1=2^{K(m)}m'$ with $m'$ odd and $K(m)=v_2(3m+1)\ge 1$; randomizing these block lengths yields
a stochastic approximation calibrated via a Dirichlet-multinomial update.
On held-out data, the NB2-GLM achieves substantially higher predictive likelihood than the odd-block
generators. Conditioning the block-length distribution on $m\bmod 8$ markedly improves the generator’s
distributional fit, indicating that low-order modular structure is a key driver of heterogeneity in
$\tau(n)$.
\end{abstract}

\section{Introduction}
\label{sec:intro}

Let $T:\NN\to\NN$ be the Collatz map
\[
T(n)=\begin{cases}
n/2, & n \text{ even},\\
3n+1, & n \text{ odd},
\end{cases}
\]
and define the total stopping time
\[
\tau(n)=\min\{t\ge 0:\ T^t(n)=1\}.
\]
The Collatz conjecture asserts $\tau(n) < \infty$ for all $n$ and remains open \cite{lagarias1985, tao2022, terras1976, sinai1992}. Here we do not attempt to prove the conjecture. Instead we treat
$n$ as random and study the induced empirical law of $\tau(n)$ over a large range.

\paragraph{Dataset and question.}
We fix $N=10^7$ and compute $\tau(n)$ for all $1\le n\le N$, yielding
$\mathcal D_N=\{(n,\tau(n))\}_{n=1}^N$. Our goal is predictive and explanatory:
\emph{which simple probabilistic models can predict and explain the observed distributional shape of
$\tau(n)$ and its arithmetic heterogeneity?}

\paragraph{Two complementary models.}
We develop:
(i) a Bayesian hierarchical Negative Binomial regression treating $\tau(n)$ as an overdispersed count whose
conditional mean depends on $\log n$ and $n\bmod 8$ (Section~\ref{sec:bayes-nb}); and
(ii) a mechanistic generative odd-block model that randomizes the block-lengths
$K=\vTwo(3m+1)$ in the accelerated odd-to-odd dynamics (Section~\ref{sec:oddblock}).
Section~\ref{sec:comparison} compares predictive performance on held-out data.
Further rationale for the regressors and the odd-block heuristics is given in Appendix~\ref{app:rationale}.

\paragraph{A note on randomness (working likelihood).}
The map $T$ and the stopping time $\tau(n)$ are deterministic.
Throughout, the term ``probabilistic model'' is used in the standard statistical sense of a
\emph{working likelihood}: we treat $n$ as random (e.g.\ uniformly sampled from $\{1,\dots,N\}$ in our evaluation),
and we model the induced variability of $\tau(n)$ across $n$ by a parametric stochastic family.
This provides uncertainty quantification and principled predictive comparisons, without positing
physical noise in the Collatz dynamics.

\section{Data and exploratory analysis}
\label{sec:data}

\subsection*{Computing $\tau(n)$ at scale}

A direct simulation for every $n\le 10^7$ is expensive due to long trajectories.
We use dynamic programming: while computing $\tau(i)$, if the orbit hits some $m<i$, we stop and set
$\tau(i)=k+\tau(m)$, reusing cached values. Numba JIT compilation accelerates the core loop
\citep{numba2015}. Details are in Appendix~\ref{app:methoddata}.

\subsection*{Distributional shape and overdispersion}

\begin{figure}[t]
\centering
\includegraphics[width=0.95\linewidth]{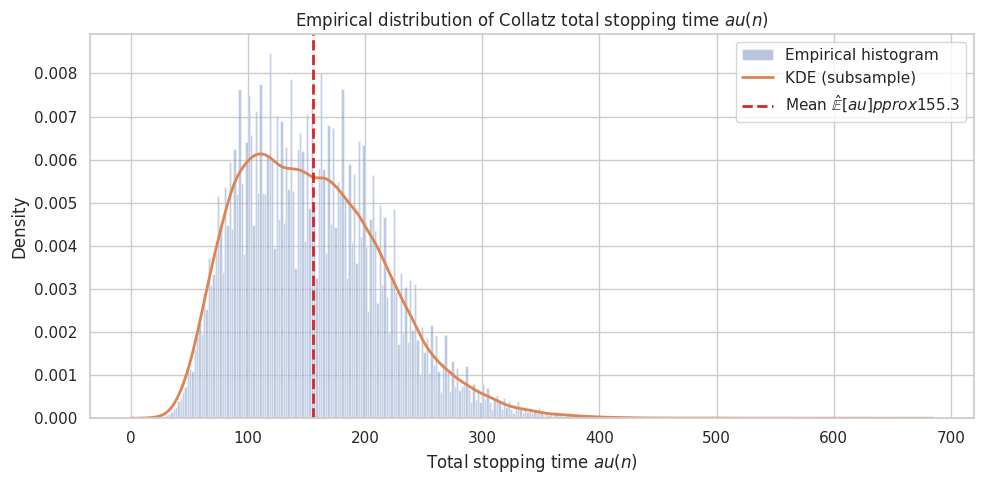}
\caption{Empirical distribution of $\tau(n)$ for $1\le n\le N$ (integer-aligned bins, width 2) with a KDE
overlay computed on a large subsample to reduce noise. This motivates an overdispersed count likelihood.}
\label{fig:histkde}
\end{figure}

Figure~\ref{fig:histkde} summarizes the marginal distribution of $\tau(n)$ (histogram with integer-aligned
bins plus KDE overlay). The heavy right tail motivates an overdispersed count likelihood.
Table~\ref{tab:tau_summary} 
 reports key summaries for $N = 10^7$, including the dispersion ratio $R=\widehat{\Var}(\tau)/\widehat{\EE}[\tau]\approx 24.56\gg 1$, ruling out Poisson as a realistic noise model \cite{cameron2013, gelman2013}.

\begin{table}[t]
\centering
\begin{tabular}{l r}
\toprule
Statistic & Value \\
\midrule
$N$ & 10{,}000{,}000 \\
$\tau_{\min}$ & 0 \\
$\tau_{\max}$ & 685 \\
$\widehat{\EE}[\tau]$ & 155.272 \\
$\widehat{\mathrm{Var}}(\tau)$ & 3814.045 \\
$\widehat{\mathrm{Var}}(\tau)/\widehat{\EE}[\tau]$ & 24.564 \\
\bottomrule
\end{tabular}
\caption{Summary statistics for $\tau(n)$ over $1\le n\le N$ .}
\label{tab:tau_summary}
\end{table}

\subsection*{Scale effect and arithmetic heterogeneity}

\begin{figure}[t]
\centering
\includegraphics[width=0.95\linewidth]{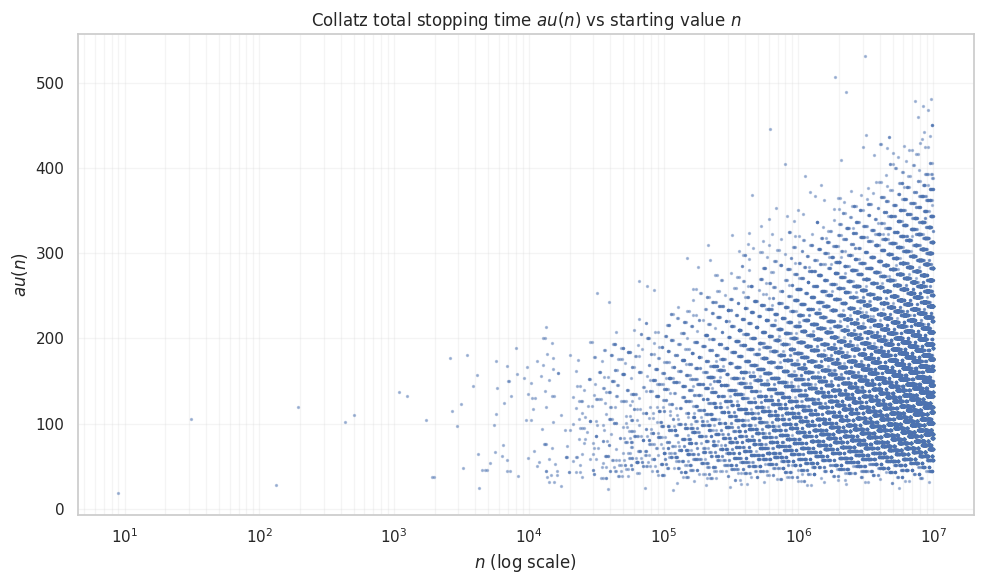}
\caption{Scatter of $\tau(n)$ vs.\ $n$ (log-$x$). The mean increases slowly and is approximately linear as a function of $\log n$,
while the spread grows with $n$; banding suggests modular structure, motivating $\log n$ and $n\bmod 8$ as covariates.}
\label{fig:scatter}
\end{figure}

Figure~\ref{fig:scatter} plots $\tau(n)$ versus $n$ on a log-$x$ axis: the mean grows slowly with $n$ and the
spread increases (heteroskedasticity). The visible banding reflects arithmetic structure; one robust proxy
is the residue class modulo a small power of two, motivating $n\bmod 8$ as a low-dimensional categorical
feature in the regression model.

\section{Method 1: Bayesian Negative Binomial regression}
\label{sec:bayes-nb}

The empirical $R\gg 1$ suggests a Negative Binomial likelihood.
We use the NB2 mean-dispersion parameterization: conditional on $(\mu_n,\alpha)$,
\[
Y_n=\tau(n)\mid \mu_n,\alpha \sim \mathrm{NB}(\mu_n,\alpha),
\qquad
\EE[Y_n\mid \mu_n,\alpha]=\mu_n,\quad
\Var(Y_n\mid \mu_n,\alpha)=\mu_n+\alpha\,\mu_n^2,
\]
with parameter mappings and a PyMC implementation note in Appendix~\ref{app:nb}.

\subsection*{Hierarchical NB2-GLM }

We link $\mu_n$ to covariates via a log-link and include a random intercept by residue class modulo 8:
\begin{align}
\log \mu_n &= \eta_n,\\
\eta_n &= \beta_0+\beta_{\log}\,\log n + u_{\mathrm{mod8}(n)},\\
u_r\mid \sigma_u &\sim \mathcal N(0,\sigma_u^2),\qquad r=0,1,\dots,7.
\end{align}
This hierarchical structure performs partial pooling across congruence classes, producing stable
class-specific deviations while controlling overfitting.

\subsection*{Inference and predictive evaluation}

We use weakly-informative priors and fully specify them in Appendix~\ref{app:nb-priors}.
We fit with NUTS \cite{hoffman2014} in PyMC \cite{pymc2023} on a training subsample of size $N_{\mathrm{fit}}=50{,}000$ and evaluate on a disjoint
test subsample of size $N_{\mathrm{test}}=50{,}000$.
Both subsamples are drawn uniformly without replacement from $\{1,\dots,N\}$ using a fixed RNG seed
(see Appendix~\ref{app:repro} for the exact protocol).
For MCMC we run $2$ chains with $1000$ tuning steps and $1000$ draws per chain, and target acceptance $0.9$.

\begin{figure}[t]
\centering
\includegraphics[width=0.90\linewidth]{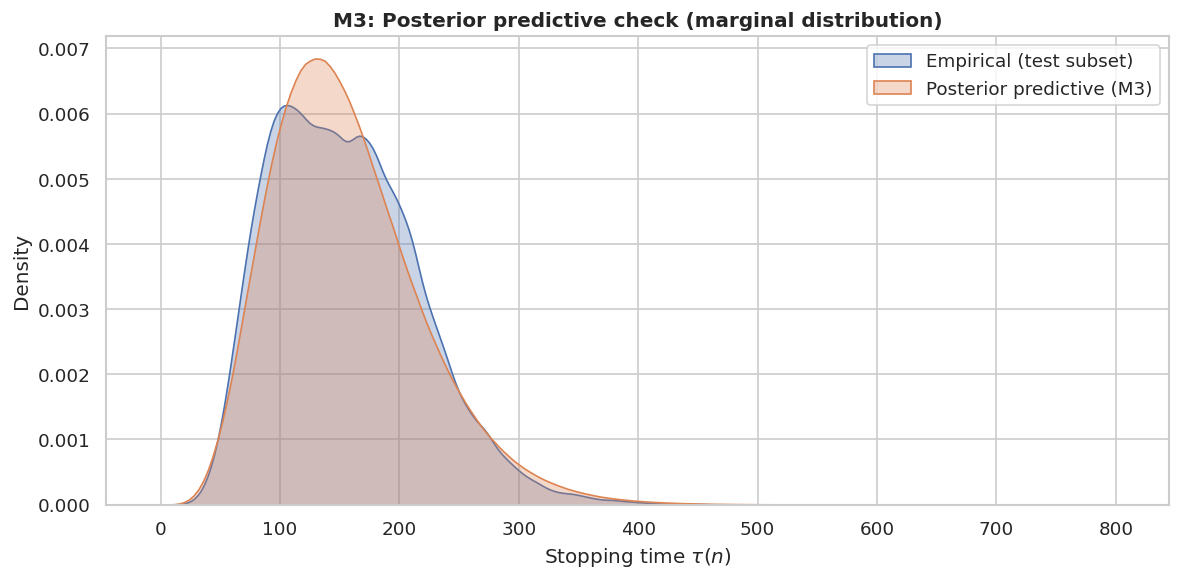}
\caption{Posterior predictive check for the hierarchical NB2-GLM (Model M3). The PPC matches the bulk well
and mildly overestimates extreme right-tail mass.}
\label{fig:ppc-m3}
\end{figure}

Figure~\ref{fig:ppc-m3} shows a held-out posterior predictive check (PPC): the model matches the bulk
distribution well and slightly inflates the far right tail, consistent with the quadratic NB2 variance.

\section{Method 2: A stochastic odd-block generative model}
\label{sec:oddblock}

Regression is predictive but phenomenological. We therefore build a mechanistic approximation based on the
odd-block decomposition. When $m$ is odd, $3m+1$ is even and can be written
\[
3m+1=2^{K(m)}m',\qquad m' \text{ odd},\qquad K(m)=\vTwo(3m+1)\ge 1.
\]
Thus one odd update is followed by $K(m)$ halving steps until returning to an odd state.

\subsection*{Accelerated odd-to-odd dynamics}

Let $m_0$ be the first odd iterate encountered from $n$ (after $\vTwo(n)$ initial halvings).
Define the accelerated map
\[
m_{j+1}=\frac{3m_j+1}{2^{K(m_j)}}.
\]
The total stopping time decomposes as
\[
\tau(n)=\vTwo(n)+\sum_{j=0}^{J-1}\bigl(1+K(m_j)\bigr),
\]
where $J$ is the first index such that $m_J=1$. This identity is exact for the deterministic dynamics.

\subsection*{Randomized block-length generator}

The generative approximation replaces $K(m_j)$ by a stochastic sequence $(K_j)$ with pmf $(p_k)_{k\ge 1}$:
\begin{equation}
\label{eq:stoch_update}
M_{j+1} \approx \roundodd{\frac{3M_j+1}{2^{K_j}}},
\qquad
\tau_{\text{gen}}(n)=\vTwo(n)+\sum_{j=0}^{J-1}(1+K_j).
\end{equation}
The rounding $\roundodd{\cdot}$ mirrors the implementation detail that the stochastic update can produce an even intermediate due to integer arithmetic, and we project back to the odd state space to preserve the odd-to-odd structure.

A classical heuristic suggests $K$ is close to geometric with $\PP(K=k)\approx 2^{-k}$, $k\ge 1$.
Under this reference, $\EE[3/2^{K}]=1$ but the log-drift is negative
(Appendix~\ref{app:logrw}), consistent with contraction on a log scale.

\paragraph{Definition of the odd projection $\roundodd{\cdot}$.}
In the generator we implement the odd-to-odd update by first rounding to the nearest integer and then
projecting to the odd state space:
let $\lfloor x \rceil$ denote rounding to the nearest integer (ties to even), and define
\[
\roundodd{x}=
\begin{cases}
1, & \lfloor x \rceil \le 1,\\
\lfloor x \rceil, & \lfloor x \rceil \text{ odd},\\
\lfloor x \rceil+1, & \lfloor x \rceil \text{ even and } \lfloor x \rceil>1.
\end{cases}
\]
This matches the implementation used to generate the results in Section~\ref{sec:comparison}.

\subsection*{Calibrating $p_k$ and posterior predictive checks}

We estimate $p_k$ from observed block lengths $K_i=\vTwo(3m_i+1)$ using a Dirichlet prior, yielding a
conjugate Dirichlet posterior \cite{murphy2012} (Appendix~\ref{app:dirichlet}).

In implementation we discretize $K$ on $\{1,\dots,K_{\max}\}$ with $K_{\max}=30$ by capping
$K_{\mathrm{cap}}=\min(K,K_{\max})$, so the last category aggregates the tail event $\{K\ge K_{\max}\}$.
We use a symmetric Dirichlet prior with concentration $\alpha_k=1$ for $k=1,\dots,K_{\max}$.

\begin{figure}[t]
\centering
\includegraphics[width=0.95\linewidth]{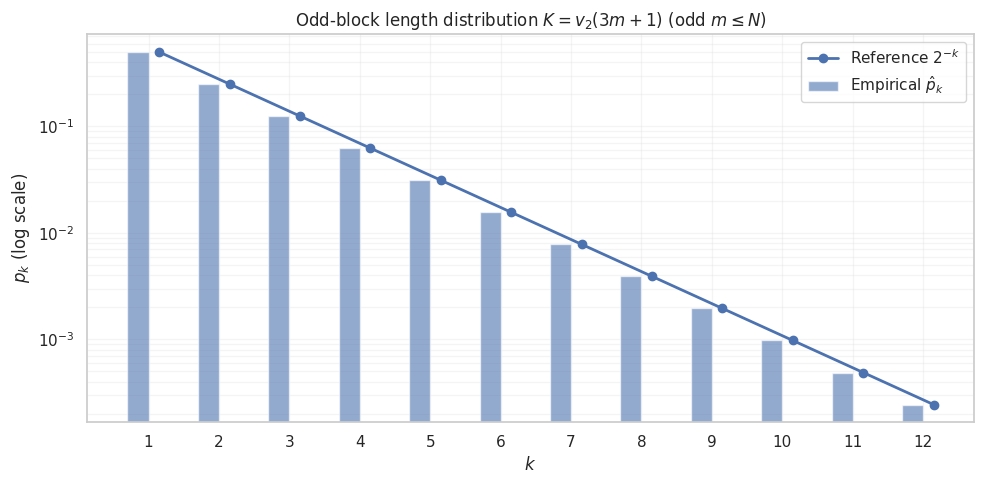}
\caption{Empirical block-length distribution $\hat p_k$ for $K=\vTwo(3m+1)$ (odd $m\le N$) vs.\ geometric
reference $2^{-k}$ on log-$y$. This evaluates the ``geometric $K$'' heuristic.}
\label{fig:pk}
\end{figure}

\begin{figure}[t]
\centering
\includegraphics[width=0.95\linewidth]{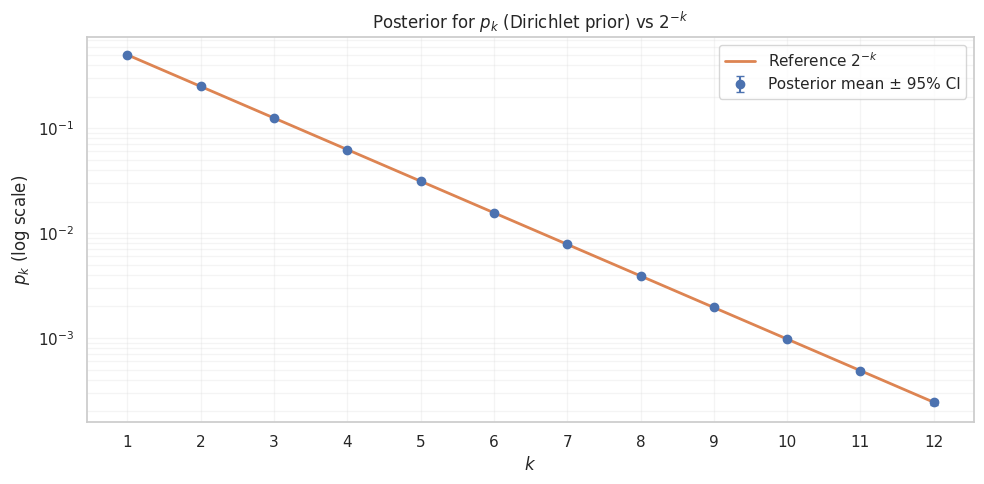}
\caption{Dirichlet posterior for $(p_k)$ (log scale) with uncertainty bars, compared to the geometric
reference $2^{-k}$.}
\label{fig:pk_post}
\end{figure}

\begin{figure}[t]
\centering
\includegraphics[width=0.95\linewidth]{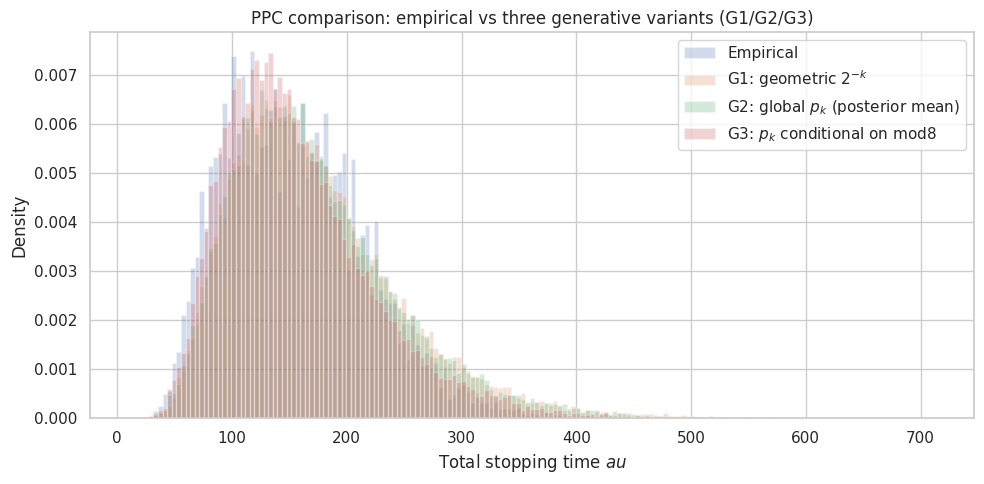}
\caption{PPC overlay: empirical $\tau$ vs.\ G1 (geometric $2^{-k}$), G2 (global calibrated $p_k$), and G3
($p_k$ conditional on $m\bmod 8$). Conditioning reduces the bias in the mean and improves distributional
agreement in the bulk.}
\label{fig:ppc_g123}
\end{figure}

Figure~\ref{fig:pk} compares empirical frequencies to the geometric reference on a log scale; this directly
tests the key modeling assumption.
Figure~\ref{fig:pk_post} shows posterior uncertainty on $(p_k)$.
Figure~\ref{fig:ppc_g123} compares the generated stopping-time distributions to empirical stopping times via a distributional PPC.

\begin{figure}[t]
\centering
\includegraphics[width=0.95\linewidth]{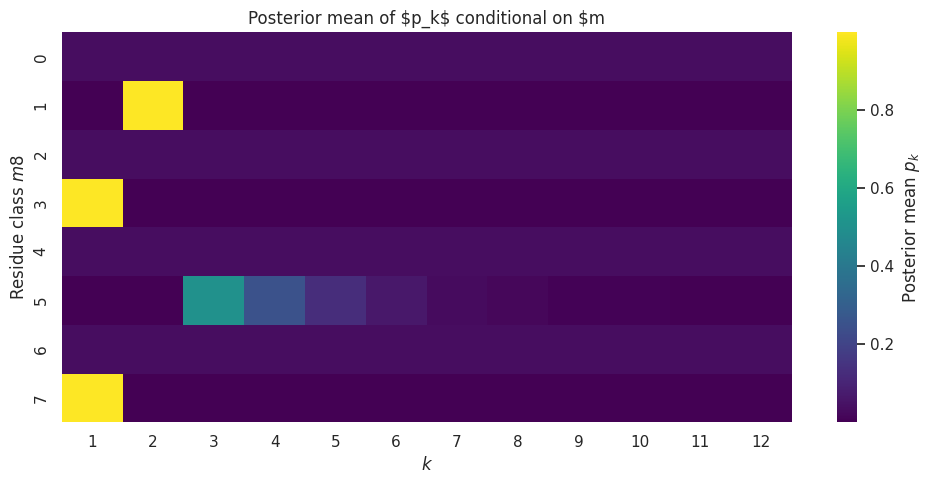}
\caption{Posterior mean of $p_k$ conditional on the odd residue class $m \bmod 8$ (heatmap). This reveals
systematic arithmetic dependence beyond an i.i.d.\ geometric law.}
\label{fig:pk_mod8}
\end{figure}

\section{Model comparison: predictive accuracy vs.\ mechanistic faithfulness}
\label{sec:comparison}

Posterior predictive checks are qualitative. To compare the regression and generative approaches on equal
footing we use a \emph{proper scoring rule} \cite{gneiting2007} on held-out data: the log predictive score. On a test set
$\{(n_i,y_i)\}_{i=1}^{N_{\mathrm{test}}}$ with $y_i=\tau(n_i)$, define
\[
\mathrm{Score}=\sum_{i=1}^{N_{\mathrm{test}}}\log p(y_i\mid n_i,\mathcal D_{\mathrm{fit}}).
\]
For the NB2-GLM, $p(y_i\mid n_i,\mathcal D_{\mathrm{fit}})$ is obtained by averaging the NB pmf over posterior
draws.
For the odd-block generator, we estimate a predictive pmf by Monte Carlo simulation:
$\hat p_i(y)=\frac1{S_{\mathrm{MC}}}\sum_{s=1}^{S_{\mathrm{MC}}} \1\{\tau_{\mathrm{gen}}^{(s)}(n_i)=y\}$ and score
$\sum_i \log(\hat p_i(y_i)+\varepsilon)$ with $\varepsilon=10^{-12}$ for numerical stability.
We use $S_{\mathrm{MC}}=40$ Monte Carlo replicates per test point.
We emphasize that the generator log score is a Monte Carlo approximation to the ideal log predictive score,
since $\hat p_i(y_i)$ is estimated from finitely many replicates.

In addition, we report a distributional distance between empirical and predictive CDFs (1-Wasserstein \cite{villani2009}) to
summarize global shape mismatch. Table~\ref{tab:comp} provides the reporting format.

\begin{table}[t]
\centering
\begin{tabular}{l c c}
\toprule
Model & Log score (higher better) & $W_1$ distance (lower better) \\
\midrule
NB2-GLM (M3)                      & $-272\,911.95$   & $3.199$ \\
Odd-block G2 (global $p_k$)       & $-1\,165\,983.43$ & $17.594$ \\
Odd-block G3 ($p_k\mid m\bmod 8$) & $-1\,079\,086.65$ & $5.434$ \\
\bottomrule
\end{tabular}
\caption{Quantitative comparison on held-out data using a proper log predictive score and a
distributional distance. All scores are computed on $N_{\mathrm{test}} = 50{,}000$ held-out pairs,
with the generative log scores estimated via $S_{\mathrm{MC}} = 40$ Monte Carlo replicates per test point.}
\label{tab:comp}
\end{table}

\paragraph{Interpretation.}
Table~\ref{tab:comp} shows that the hierarchical NB2-GLM achieves by far the best
predictive likelihood on the held-out set: its log score is
$-2.73\times 10^5$ (average $-5.46$ per observation), compared to
$-1.17\times 10^6$ for the global odd-block generator G2 and
$-1.08\times 10^6$ for the conditional generator G3.
In other words, the regression model assigns substantially higher
probability to the realized stopping times.
In terms of distributional fit, the NB2-GLM also attains the smallest
$W_1$ distance ($W_1 \approx 3.20$), with G3 improving markedly over G2
($W_1 \approx 5.43$ vs.\ $17.59$) but still lagging behind the regression.

From a purely predictive, proper-score perspective, the NB2-GLM (M3) is therefore
the best model among those considered.
However, the odd-block generator remains attractive from a mechanistic
point of view: it explains heavy tails and heterogeneity through
randomized block lengths $K = v_2(3m+1)$ and their arithmetic dependence
on $m \bmod 8$.
The conditional variant G3 narrows the gap in predictive performance
while preserving this structural interpretation, making it a natural
candidate when mechanistic faithfulness is prioritized over raw log score.

\section{Discussion}
\label{sec:discussion}

Two complementary conclusions emerge.
First, a compact hierarchical NB2-GLM with $\log n$ and $n\bmod 8$ captures much of the predictive structure
of $\tau(n)$, with uncertainty quantification via the posterior and PPC
(Figure~\ref{fig:ppc-m3}). Second, the odd-block decomposition yields a mechanistic generator whose success
hinges on accurately modeling the distribution of $K=\vTwo(3m+1)$ and its arithmetic dependence
(Figures~\ref{fig:pk}--\ref{fig:ppc_g123}). The conditional variant provides a bridge between the two
viewpoints: modular structure that appears as a random effect in regression becomes explicit conditioning
information in the generator.

Future work includes extending the conditional structure to higher powers of two, incorporating explicit
state-dependence in $(K_j)$, and using likelihood-based calibration of the generator to align scoring-rule
performance with mechanistic interpretability.

\bibliographystyle{unsrtnat}
\bibliography{collatz_bib}

\newpage
\appendix
\section*{Appendix}
\addcontentsline{toc}{section}{Appendix}

\section{Method to construct the dataset}
\label{app:methoddata}

A naive implementation of the stopping-time computation is too slow when
$N$ is as large as $10^7$, due to the potentially long trajectories required
for some starting values. To obtain a scalable engine, we use Python together
with NumPy arrays and Numba's just-in-time (JIT) compilation
\citep{numba2015}. Numba compiles the core loops to efficient machine
code, drastically reducing Python overhead and enabling the computation of
$\tau(n)$ for all $n\le N$ within a standard Colab-like runtime.

We also exploit a dynamic-programming shortcut. When computing $\tau(i)$ for a
given $i$, the trajectory $(n_k)$ may eventually visit a value $m<i$.
At that point the remaining number of steps to reach $1$ is exactly $\tau(m)$,
which has already been computed earlier. Thus we can terminate the iteration
and set
\[
\tau(i)=k+\tau(m),
\]
reusing cached values. This reuse is particularly effective for large $N$,
where many trajectories merge into segments already explored.

The output of this stage is the dataset
\[
\mathcal{D}_N=\{(n,\tau(n)):\ 1\le n\le N\},
\]
stored as a table with $N$ rows. In later analysis we augment each row with
simple derived features of $n$ (e.g.\ $\log n$ and $n\bmod 8$) to serve as
covariates in the regression model.

\section{Reproducibility details (splits, priors, and evaluation)}
\label{app:repro}

\subsection{Train/test protocol}
We draw $N_{\mathrm{fit}}=50{,}000$ indices uniformly without replacement from $\{1,\dots,N\}$ using RNG seed $123$,
and set the remaining pool as candidates for testing. We then draw $N_{\mathrm{test}}=50{,}000$ test indices uniformly
without replacement from the remaining pool, again using the same RNG stream.

\subsection{Priors and MCMC settings for M3}
\label{app:nb-priors}
For the hierarchical NB2-GLM (M3) we use:
\[
\beta_0 \sim \mathcal N(0,10^2),\qquad
\beta_{\log}\sim \mathcal N(0,5^2),\qquad
\sigma_u \sim \mathrm{HalfNormal}(1),\qquad
\alpha \sim \mathrm{HalfNormal}(5),
\]
with $u_r=\sigma_u z_r$, $z_r\sim\mathcal N(0,1)$ for $r=0,\dots,7$.
We sample with NUTS (PyMC) using $2$ chains, $1000$ tuning steps and $1000$ posterior draws per chain,
target acceptance $0.9$, and a fixed random seed.

\subsection{Generator evaluation settings}
For the odd-block generators we cap the simulated stopping-time loop at $\texttt{MAX\_STEPS}=200{,}000$ steps to avoid
pathological non-absorption under the stochastic approximation.
Log scores use $S_{\mathrm{MC}}=40$ Monte Carlo replicates per test point, $\varepsilon=10^{-12}$, and fixed RNG seeds
for exact reproducibility.

\section{Modeling choices and heuristics (rationale)}
\label{app:rationale}

This appendix explains the main modeling choices in the paper: (i) the regressors used in the NB2-GLM,
(ii) the Negative Binomial likelihood, and (iii) the design of the odd-block generator and its variants.
The goal is not to add new theory, but to make the modeling decisions transparent and defensible.

\subsection{Why model $\tau(n)$ as an overdispersed count (NB2 rather than Poisson)}
\label{app:rationale_nb}

The response $\tau(n)$ is a nonnegative integer-valued quantity, so a count likelihood is natural.
A Poisson model would impose $\Var(Y\mid x)\approx \E[Y\mid x]$, which is strongly violated in our data:
the empirical dispersion ratio $\widehat{\Var}(\tau)/\widehat{\E}[\tau]$ is far above $1$
(Table~\ref{tab:tau_summary}). This motivates a Negative Binomial likelihood, which allows variance to grow
faster than the mean.

We use the NB2 mean--dispersion parameterization
\[
\Var(Y\mid \mu,\alpha)=\mu+\alpha\mu^2,
\]
because it captures two empirically visible features simultaneously:
(i) strong overdispersion and (ii) increasing spread at larger conditional means (heteroskedasticity).
In practice, NB2 is also a standard default in applied count regression when the data show heavy tails and
variance inflation.

\subsection{Why $\log n$ as the main scale regressor}
\label{app:rationale_logn}

A basic empirical fact is that $\tau(n)$ grows slowly with $n$ but not linearly (Figure~\ref{fig:scatter}).
Using $n$ directly as a regressor would force the model to allocate too much curvature to match a trend that
is visually closer to logarithmic.
The transformation $\log n$ provides a simple one-dimensional scale summary that:
(i) compresses the dynamic range $1\le n\le 10^7$,
(ii) matches the observed slow growth of typical stopping times,
and (iii) yields stable numerical inference under a log-link.

In short, $\log n$ is the minimal scale feature that captures the macro-trend without introducing a flexible nonlinear basis that would be harder to interpret or justify, allowing us to maintain model parsimony and interpretability.

\subsection{Why include residue class $n\bmod 8$ (and why as a random effect)}
\label{app:rationale_mod8}

Beyond scale effects, Collatz dynamics exhibit visible arithmetic stratification:
values of $\tau(n)$ form bands when plotted against $n$ (Figure~\ref{fig:scatter}),
indicating that a coarse modular signature already explains a nontrivial part of the heterogeneity.
We use $n\bmod 8$ as a low-dimensional categorical covariate because:
\begin{itemize}
\item It is a minimal ``power-of-two'' modular feature (small enough to avoid a high-dimensional categorical design),
\item It aligns with the prominence of 2-adic structure in Collatz steps (repeated halving), and
\item It is empirically robust: the banding is clearly visible already at this resolution.
\end{itemize}

We model the eight class offsets as a hierarchical (partially pooled) random effect
$u_r\sim\mathcal N(0,\sigma_u^2)$ rather than eight unrelated fixed effects.
This choice stabilizes inference and prevents overfitting by shrinking class-specific deviations toward $0$
when the data provide weak evidence for large differences.
It also matches the modeling goal: we want to quantify that ``mod 8 matters'' while keeping the model small.

\subsection{Why we did not add more regressors }
\label{app:rationale_nomore}

Additional predictors are possible (e.g.\ $v_2(n)$, richer residue classes, or nonlinear functions of $\log n$).
We deliberately keep the regression model compact to preserve interpretability: the point is not to maximize predictive accuracy at any cost, but to demonstrate that
a minimal set of theoretically motivated covariates already yields strong predictive performance and useful
uncertainty quantification.

\subsection{Odd-block generator: why it is a natural mechanistic approximation}
\label{app:rationale_oddblock}

The odd-block decomposition isolates the only ``expanding'' step (odd $\mapsto 3m+1$) from the subsequent
sequence of deterministic halving steps.
For odd $m$, writing $3m+1=2^{K(m)}m'$ with $m'$ odd turns the dynamics into odd-to-odd jumps plus a
block length $K(m)$.
This motivates a mechanistic approximation where the complicated deterministic dependence of $K(m)$ on $m$
is replaced by a stochastic surrogate distribution.

The baseline heuristic $\Pr(K=k)\approx 2^{-k}$ corresponds to the intuition that $3m+1$ behaves roughly like
a ``random even integer'' in its 2-adic valuation, so the number of factors of $2$ is approximately geometric.
The calibrated variants (G2/G3) keep the same mechanism but estimate the block-length distribution from data,
making the approximation testable rather than purely heuristic.

\subsection{Why conditioning on $m\bmod 8$ is the first meaningful refinement}
\label{app:rationale_g3}

The global i.i.d.\ block-length model (G2) averages over strong arithmetic heterogeneity.
A minimal refinement is to condition $p_k$ on a low-order residue class that is directly tied to 2-adic
structure. Conditioning on $m\bmod 8$ is the smallest such choice that is still informative:
it is low-dimensional (8 classes) but already captures systematic differences in the estimated block-length
distribution (Figure~\ref{fig:pk_mod8}). The improved PPC and reduced $W_1$ distance for G3 indicate that the
generator benefits substantially from this modular information.

\subsection{Why these evaluation metrics}
\label{app:rationale_metrics}

We report two complementary diagnostics:
\begin{itemize}
\item \textbf{Log predictive score} (proper scoring rule): it evaluates the entire predictive distribution,
rewarding models that assign higher probability to the observed outcomes.
This is the main principled metric for predictive comparison.
\item \textbf{$W_1$ (1-Wasserstein) distance}: it summarizes global distributional mismatch between the empirical
and predictive marginals, which is particularly informative when heavy tails are present and visual PPCs show
shape differences.
\end{itemize}

For the odd-block generator, the predictive distribution is not available in closed form, so the log score is
estimated by Monte Carlo ``hit'' frequencies at each test point, with $S_{\mathrm{MC}}$ replicates.
This yields an approximation to a proper score; increasing $S_{\mathrm{MC}}$ tightens the estimate at higher
computational cost.

\section{Negative Binomial regression: parameterizations and properties}
\label{app:nb}

This appendix summarizes the Negative Binomial (NB) distribution and the NB
generalized linear model (GLM) conventions used in the main text.

\subsection{Classical $(r,p)$ parameterization}

The classical Negative Binomial distribution can be defined as the
distribution of the number of failures $Y$ observed before the $r$-th
success in i.i.d.\ Bernoulli trials with success probability $p\in(0,1)$:
\[
Y\sim \mathrm{NB}(r,p),\qquad r>0,\; p\in(0,1).
\]
Its pmf is
\[
\Pr(Y=y\mid r,p)=
\frac{\Gamma(y+r)}{\Gamma(r)\,\Gamma(y+1)}(1-p)^{y}p^{r},
\qquad y=0,1,2,\dots,
\]
and
\[
\EE[Y]=r\,\frac{1-p}{p},\qquad \Var(Y)=r\,\frac{1-p}{p^2}.
\]

\subsection{Mean--dispersion $(\mu,\alpha)$ parameterization (NB2)}

For regression we use the NB2 convention
\[
\EE[Y\mid \mu,\alpha]=\mu,\qquad
\Var(Y\mid \mu,\alpha)=\mu+\alpha\,\mu^2,
\]
where $\alpha>0$ controls overdispersion and $\alpha\to 0$ yields the Poisson limit.
The mapping to $(r,p)$ is
\[
r=\frac{1}{\alpha},\qquad p=\frac{1}{1+\alpha\mu}.
\]

\subsection{Implementation note: mapping to PyMC}

In PyMC, \texttt{NegativeBinomial(mu=mu, alpha=alpha\_PyMC)} uses
\[
\Var(Y\mid\mu,\alpha_{\mathrm{PyMC}})=\mu+\frac{\mu^2}{\alpha_{\mathrm{PyMC}}}.
\]
Therefore, to match the NB2 convention above we set
\[
\alpha_{\mathrm{PyMC}}=\frac{1}{\alpha}.
\]

\subsection{Poisson--Gamma mixture view}

If $\Lambda\sim\mathrm{Gamma}(r,c)$ and $Y\mid\Lambda\sim\mathrm{Poisson}(\Lambda)$, then
$Y\sim\mathrm{NB}(r,p)$ with $p=c/(c+1)$. In NB2 form, taking $r=\alpha^{-1}$ and $c=\alpha^{-1}/\mu$
recovers $\Var(Y)=\mu+\alpha\mu^2$, interpreting overdispersion as latent rate heterogeneity.

\section{Dirichlet--multinomial update}
\label{app:dirichlet}

If $K_1,\dots,K_m$ are i.i.d.\ categorical with probabilities $p_1,\dots,p_{K_{\max}}$ and prior
$p\sim\mathrm{Dir}(\alpha)$, the posterior is $\mathrm{Dir}(\alpha+c)$ with
$c_k=\sum_{i=1}^m \1\{K_i=k\}$. Posterior means and variances are
\[
\EE[p_k\mid K_{1:m}] = \frac{\alpha_k+c_k}{\sum_{j}(\alpha_j+c_j)}, \qquad
\mathrm{Var}(p_k\mid K_{1:m}) =
\frac{(\alpha_k+c_k)\left(\alpha_0+m-(\alpha_k+c_k)\right)}{(\alpha_0+m)^2(\alpha_0+m+1)},
\]
where $\alpha_0=\sum_j \alpha_j$.

\section{Log-scale random walk approximation}
\label{app:logrw}

Ignoring the additive $+1$ term for large states, the odd-block update satisfies
$M_{j+1}\approx (3/2^{K_j}) M_j$, hence in logarithmic scale
\[
X_{j+1} \approx X_j + \log_2 3 - K_j, \qquad X_j=\log_2 M_j.
\]
Even when $\EE[3/2^{K}]=1$, Jensen’s inequality implies
$\EE[\log_2(3/2^{K})]<\log_2 \EE[3/2^{K}]=0$,
yielding a negative drift in $X_j$ and suggesting absorption at small values is typical.


\begin{figure}[t]
\centering
\includegraphics[width=0.95\linewidth]{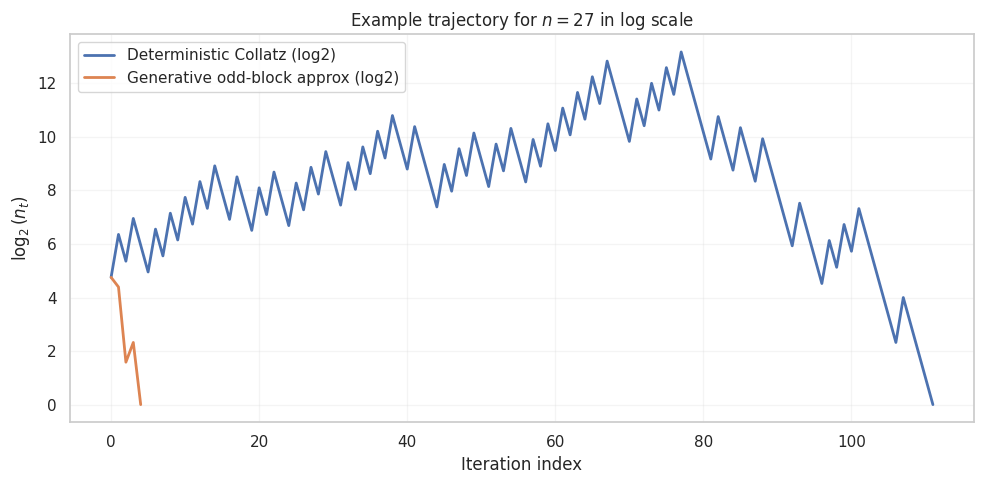}
\caption{Example trajectory in log scale for $n=27$: deterministic Collatz vs.\ stochastic odd-block approximation.}
\label{fig:traj}
\end{figure}


\end{document}